\documentclass[runningheads]{llncs}

\usepackage{todonotes}
\usepackage[title]{appendix}
\usepackage{caption}
\usepackage{float}
\usepackage{amsmath}
\usepackage{amssymb}
\usepackage{subcaption}
\usepackage{color,soul}
\usepackage[normalem]{ulem}

\usepackage{booktabs}

\begin{document}
\title{Disentangling What and Where for 3D Object-Centric Representations Through Active Inference}
\titlerunning{What and Where through Active Inference}
\author{Toon Van de Maele, Tim Verbelen, Ozan \c{C}atal and Bart Dhoedt}
\authorrunning{T. Van de Maele et al.}
\institute{
    IDLab, Department of Information Technology\\
    Ghent University - imec\\
    Ghent, Belgium\\
    \email{firstname.lastname@ugent.be}
}

\maketitle              


\begin{abstract}
    Although modern object detection and classification models achieve high accuracy, these are typically constrained in advance on a fixed train set and are therefore not flexible to deal with novel, unseen object categories. Moreover, these models most often operate on a single frame, which may yield incorrect classifications in case of ambiguous viewpoints. In this paper, we propose an active inference agent that actively gathers evidence for object classifications, and can learn novel object categories over time. Drawing inspiration from the human brain, we build object-centric generative models composed of two information streams, a what- and a where-stream. The what-stream predicts whether the observed object belongs to a specific category, while the where-stream is responsible for representing the object in its internal 3D reference frame. We show that our agent (i) is able to learn representations for many object categories in an unsupervised way, (ii) achieves state-of-the-art classification accuracies, actively resolving ambiguity when required and (iii) identifies novel object categories. Furthermore, we validate our system in an end-to-end fashion where the agent is able to search for an object at a given pose from a pixel-based rendering. We believe that this is a first step towards building modular, intelligent systems that can be used for a wide range of tasks involving three dimensional objects.
\end{abstract}

\section{Introduction}
    
    
    In the last decade, we have seen a proliferation of deep learning systems, especially in the field of image classificaton~\cite{alexnet,resnets}. Although these systems show high accuracies on various classification benchmarks, their applicability is typically limited to a fixed input distribution based on the dataset used during training. In contrast, the real world is not stationary, which urges the need for continual learning~\cite{Hadsell2020}. Also, these classifiers lack the concept of action, which renders them vulnerable to ambiguous and adverserial samples~\cite{adversarial}. As humans, we will typically move around and sample more viewpoints to improve the precision of our classification, illustrating the importance of embodiment in building intelligent agents~\cite{embodied}.
    
    Active inference offers a unified treatment of perception, action and learning, which states that intelligent systems build a generative model of their world and operate by minimizing a bound on surprise, i.e. the variational free energy~\cite{Friston2016}. In~\cite{parr2021activevision}, Parr et al. propose a model for (human) vision, which considers a scene as a factorization of separate (parts of) objects, encoding their identity, scale and pose. This is in line with the so called two stream hypothesis, which states that visual information is processed by a dorsal (``where'') stream on the one hand, representing where an object is in the space, and a ventral (``what'') stream on the other hand, representing object identity~\cite{whatwhere}. Similarly, Hawkins et al. propose that cortical columns in the neocortex track objects and their pose in a local reference frame, encoded by cortical grid cells~\cite{Hawkings2019Cortical}.
    
    In this paper, we propose a system that builds on these principles for learning object-centric representations that allow for accurate classification. Inspired by cortical columns, our system is composed of separate deep neural networks, called Cortical Column Networks (CCN), where each CCN learns a representation of a single type of 3D object in a local reference frame. The ensemble of CCNs forms the agent's generative model, which is optimized by minimizing free energy. By also minimizing the expected free energy in the future, we show that our agent can realize preferred viewpoints for certain objects, while also being urged to resolve ambiguity on object identiy.
    
    We evaluate our agent on pixel data rendered from 3D objects from the YCB benchmarking dataset~\cite{ycb}, where the agent can control the viewpoint. We compare the performance of an embodied and a static agent for classification, and show that classification accuracy is higher for the embodied agent. Additionally, we leverage the where stream for implicit pose estimation of the objects. 

\section{Method} 
    \label{sect:method} 
    
    
    In active inference, an agent acts and learns in order to minimize an upper bound on the negative log evidence of its observations, given its generative model of the world i.e. the free energy. In this section, we first formally introduce the generative model of our agent for representing 3D objects. Next we discuss how we instantiate and train this generative model using deep neural networks. Finally, we show how action selection is driven by minimizing expected free energy in the future.
    
    \subsection{A generative model for object-centric perception} 

    Our generative model is based on~\cite{parr2021activevision}, but focused on representing a single object. Concretely, our agent obtains pixel observations $\mathbf{o}_{0:t}$ that render a 3D object with identity $\mathbf{i}$ as viewed from certain viewpoints $\mathbf{v}_{0:t}$ specified in an object-local reference frame. Each time step $t$ the agent can perform an action $\mathbf{a}_t$, resulting in a relative translation and rotation of the camera. The joint probability distribution then factorizes as:
    
    \begin{equation}
        p(\mathbf{o}_{0:t},\mathbf{a}_{0:t-1},\mathbf{v}_{0:t}, \mathbf{i}) = p(\mathbf{i}) \prod_{t} p(\mathbf{o}_t|\mathbf{v}_t, \mathbf{i}) p(\mathbf{v}_t | \mathbf{v}_{t-1}, \mathbf{a}_{t-1}) p(\mathbf{a}_{t-1}) 
        \label{eq:generativemodel}
    \end{equation}
    
    Using the approximate posterior $q(\mathbf{i}, \mathbf{v}_{0:t}|\mathbf{o}_{0:t}) = q(\mathbf{i}|\mathbf{o}_{0:t})\sum_{t}q(\mathbf{v}_t|\mathbf{i}, \mathbf{o}_t)$, the free energy becomes:
    
    \begin{equation}
    \begin{split}
        F &= \mathbb{E}_{q(\mathbf{i}, \mathbf{v}_{0:t})}[ \log q(\mathbf{i}, \mathbf{v}_{0:t}|\mathbf{o}_{0:t}) - \log p(\mathbf{o}_{0:t},\mathbf{a}_{0:t-1},\mathbf{v}_{0:t}, \mathbf{i})] \\
        &\stackrel{+}{=} D_{KL}[q(\mathbf{i}|\mathbf{o}_{0:t})||p(\mathbf{i})] + \sum_{t} D_{KL}[q(\mathbf{v}_t|\mathbf{i}, \mathbf{o}_t) || p(\mathbf{v}_t|\mathbf{v}_{t-1}, \mathbf{a}_{t-1})] \\
        & \;\;\;\; - \mathbb{E}_{q(\mathbf{i}, \mathbf{v}_{0:t})}[\log p(\mathbf{o}_t|\mathbf{v}_t, \mathbf{i})]
    \end{split}
    \label{eq:fe} 
    \end{equation}
    
    This shows that minimizing free energy is equivalent to maximizing the accuracy, i.e. predicting the observation for a given object identity and viewpoint, while minimizing complexity of the posterior models.

    \subsection{An ensemble of CCNs}

    We instantiate the generative model using deep neural networks similar to a variational autoencoder (VAE)~\cite{kingma2013auto,rezende2014stochastic} with an encoder and decoder part. For each object identity, we train a separate encoder-decoder pair, since $p(\mathbf{o}_t|\mathbf{v}_t, \mathbf{i}) = \sum_k p(\mathbf{o}_t|\mathbf{v}_t, i=k)$. Similarly the encoder outputs distribution parameters for the object identity $q(i=k|\mathbf{o}_t)$ and viewpoint $q(\mathbf{v}_t|\mathbf{o}_t, i=k)$, the former parameterized as a Bernoulli variable, the latter as a multivariate Gaussian with a diagonal covariance matrix. Finally, we also parameterize the transition model $p(\mathbf{v}_t|\mathbf{v}_{t-1}, \mathbf{a}_{t-1}, i=k)$ which enforces $\mathbf{v}$ to encode relative viewpoint information.
    
\begin{figure}[ht]
      \centering
      \includegraphics[width=0.8\linewidth]{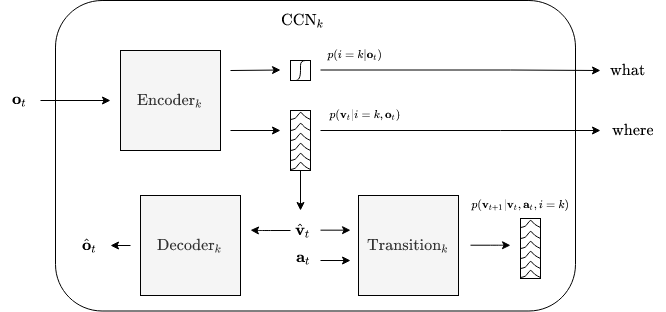}  
      \caption{A single CCN. Observation $\mathbf{o}_t$ is processed by an encoder and provides a belief over the object identity $p(i=k|\mathbf{o}_t)$ and over the observers pose  $p(\mathbf{v}_t|i=k, \mathbf{o}_t)$. From this distribution, a sample $\mathbf{\hat{v}}_t$ is drawn and is decoded in a reconstructed observation $\mathbf{\hat{o}}_t$. This sample is also transformed together with an action $\mathbf{a}_t$ into a belief over a future pose $\mathbf{v}_{t+1}$.}
      \label{fig:ccna}
    \end{figure}

    Intuitively, each encoder-decoder pair captures the information about a single object class, with a ``what'' stream modeled as a binary classifier of whether an observation belongs to a certain object identiy, and a ``where'' stream encoding the observer viewpoint w.r.t. a local, object-specific reference frame. We call such a pair a Cortical Column Network (CCN), as it mimicks the ``voting for object at pose'' behavior of cortical columns in the neocortex as hypothesized in~\cite{Hawkings2019Cortical}. This is illustrated in Figure~\ref{fig:ccna}.
    The agent hence entails a generative model as an ensemble of CCNs. We obtain $q(\mathbf{i}|\mathbf{o}_{0:t}) \propto q(\mathbf{i}|\mathbf{o}_{0:t-1})q(\mathbf{i}|\mathbf{o}_{t})$, where $q(\mathbf{i}|\mathbf{o}_{t})$ is a Categorical distribution from the CCN votes $q(i=k|\mathbf{o}_t)$, and $q(\mathbf{i}|\mathbf{o}_{0:t-1})$ is a conjugate prior Dirichlet distribution whose concencentration parameters are aggregated votes from previous observations, as updated in a Bayesian filter~\cite{activeinferencetutorial}. This process computes the posterior belief over the different timesteps. The Dirichlet distribution reflects the prior that an object is unlikely to change its category between timesteps. We also include an ``other'' object class, which is activate when none of the object classes receive votes, hence enabling the agent to detect novel object categories.
    
    Each CCN is trained in an end-to-end fashion using a dataset of object observation pairs and the relative camera transform between them for each object class. To minimize Equation~\ref{eq:fe}, we use MSE loss on the reconstructions and a KL divergence between the viewpoint posterior and transition model. The identity posterior is trained as a binary classifier, sampling positive and negative anchors from the dataset. For more details on the training loss and model architectures, the reader is referred to the appendix.

    \subsection{Classification by minimizing expected free energy}

    Crucially in active inference, an agent will select the action that minimizes the expected free energy in the future $G$. In our case, this yields:
    
    \begin{equation}
    \begin{split}
        G(\mathbf{a}_t) &= \mathbb{E}_{q(\mathbf{i}, \mathbf{v}_{0:t+1},\mathbf{o}_{t+1})}[ \log q(\mathbf{i}, \mathbf{v}_{0:t+1}|\mathbf{o}_{0:t}, \mathbf{a}_{t}) - \log p(\mathbf{o}_{0:t+1},\mathbf{a}_{0:t-1},\mathbf{v}_{0:t+1}, \mathbf{i}| \mathbf{a}_{t})] \\
        &\approx \mathbb{E}_{q(\mathbf{o}_{t+1})}[- \log p(\mathbf{o}_{0:t+1})] \\ 
        & \;\;\;\; - \mathbb{E}_{q(\mathbf{i}, \mathbf{v}_{0:t+1},\mathbf{o}_{t+1})}[
        \log q(\mathbf{i}|\mathbf{o}_{0:t+1}, \mathbf{a}_t) - \log q(\mathbf{i}|\mathbf{o}_{0:t}, \mathbf{a}_t)] \\
        & \;\;\;\; - \mathbb{E}_{q(\mathbf{i}, \mathbf{v}_{0:t+1},\mathbf{o}_{t+1})}[ \log q(\mathbf{v}_{0:t+1}|\mathbf{i}, \mathbf{o}_{0:t+1} ,\mathbf{a}_t) - \log q(\mathbf{v}_{0:t+1}|\mathbf{i}, \mathbf{o}_{0:t}, \mathbf{a}_t)] 
    \raisetag{1.01\normalbaselineskip}
    \end{split}
    \label{eq:efe} 
    \end{equation}

    The expected free energy unpacks into three terms, the first is an instrumental term that indicates that the agent is driven to some prior preferred observations, whereas the second and third term encode the expected information gain for the object identity and the object pose for a certain action. This shows how the agent can be steered to seeing a certain object at a certain pose, which could be for example a grasp position in the case of a robotic manipulator. On the other hand, in the absence of preferences, the agent will query new viewpoints that provide information on the object identity and pose, effectively trying to get a better classification.

\section{Experiments} 

    We evaluate our model for an agent in a 3D environment, where 3D models of objects from the YCB dataset~\cite{ycb} are rendered from a certain camera viewpoint. The agent actions are then defined as relative transforms (i.e. rotation and translation), moving the camera viewpoint. This setup closely mimicks a robot manipulator with an in-hand camera, but without kinematic constraints~\cite{VanDeMaele}.
    
    We create a dataset using 3D meshes of objects from the YCB dataset~\cite{ycb}. For each of 9 ``known'' objects, 14000 viewpoints and their corresponding view, for which the object is centered in view, are generated as a train set. During training, pairs of two views are randomly selected, for which the action is defined as the relative transform between these two viewpoints.
    
    
    We first validate that the CCN ensemble is able to learn pose and identity representations unsupervisedly by minimizing free energy. Next, we show how the expected free energy allows to agent to infer actions that can bring the agent to a preferred pose relative to an object on the one hand, and resolve ambiguity for inferring an object identity on the other hand.
    
    \begin{figure}[b!]
        \begin{subfigure}[t]{0.48\textwidth}
            \centering
            \includegraphics[width=\textwidth]{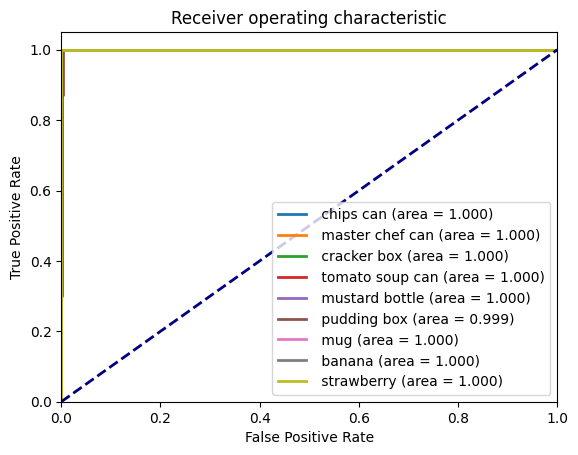}
            \caption{}
            \label{fig:roc}
        \end{subfigure}
        \hfill
        \begin{subfigure}[t]{0.48\textwidth}
            \centering
            \includegraphics[width=0.235\textwidth]{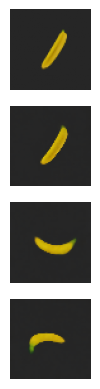}
            \includegraphics[width=0.235\textwidth]{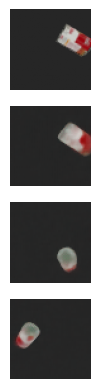}
            \includegraphics[width=0.235\textwidth]{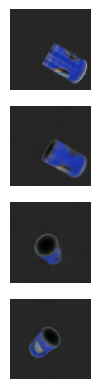}
            \includegraphics[width=0.235\textwidth]{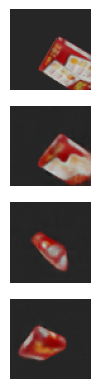}
            \caption{}
            \label{fig:generation}
        \end{subfigure}
        \caption{(a) The ROC curve for individual CCNs. Negative samples are observations from the test set: padlock, power drill, knife, orange and tuna fish can. (b) The ground truth (top row), the reconstruction (second row), and imagined transformed observations (other rows) are shown for multiple YCB objects~\cite{ycb}.}
    \end{figure}
    
    \subsection{The ``what'' stream: object recognition}
    
    
    First, we evaluate the performance of each individual CCN ``what'' binary classifier. The ROC curves are shown in Figure~\ref{fig:roc} where each CCN is tested on a dataset with 3000 novel views for each of the 9 known objects, and 3000 views from 5 objects, it has never seen during training. For all objects we achieve near-perfect ROC curves, which can be attributed to the fact that each CCN can focus on particular features that distinguish a particular object from the others. Investigating the impact on the ROC performance when using real-world observations instead of 3D renders of predefined object models would be an interesting avenue for future work.


    \subsection{The ``where'' stream: implicit pose estimation}  
    
    Crucially, our CCNs not only learn a classification output, but also an implicit representation of the 3D structure of the object at hand. As discussed in Section~\ref{sect:method}, this is encoded in a latent code $\mathbf{v}_{t}$, from which the model can reconstruct the given viewpoint using the decoder, or imagine other viewpoints after a relative transform using the transition model. This is illustrated in Figure~\ref{fig:generation}, where the first row shows ground truth object observations, the second row shows the reconstruction after encoding, and the third and fourth row show imagined other viewpoints.
    
\begin{figure}[b!]
        \centering
        \includegraphics[width=0.75\textwidth]{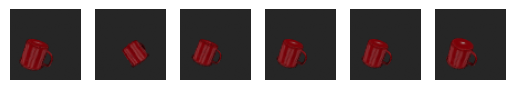}
        \includegraphics[width=0.75\textwidth]{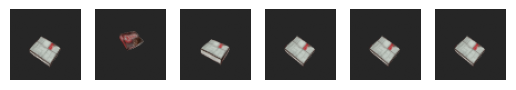}
        \caption{The agent is provided with a preferred observation (first column left), and an intial observation (second column). The agent infers the relative transforms to reach the preferred observation.}
        \label{fig:poseestimation}
    \end{figure}

    
    We can now use the CCN to infer the actions that will yield some ``preferred'' observation, by minimizing the expected free energy in Equation~\ref{eq:efe}. This is useful for example to instruct a robotic manipulator to a certain grasp point for an object. As computing $G$ for every action is intractable, we sample 1000 random relative transforms for which $G$ is calculated. A transform is sampled by first sampling a target viewpoint in 3D space uniformly in the workspace. The orientation is then determined so that the camera looks at the center of gravity of the object. The relative transform can then be computed between both current and target sampled poses. The identity transform is always provided as an option, allowing the agent to stay at its current pose when no better option is found. This results in the agent finding the estimated pose. Figure~\ref{fig:poseestimation} shows qualitative trajectories for estimating the correct pose from both a mug and a pudding box. On average, the pose estimation process converges after 3 steps, and the resulting final pose lies around 1 mm (average of 1.4 mm) and 5 degrees (average of 4.7 degree) in distance and angle compared to the ground truth. We provide a more detailed table in Appendix B. 
    
    \subsection{Embodied agents for improved classification}
    
    Whereas previously we evaluated the binary classfication performance of individual CCNs, we now evaluate the performance of the CCN ensemble as an $n+1$-way classifier, with $n$ object classes and one ``other'' class. 
    
    

    We evaluate an embodied agent that can query extra viewpoints to improve its classification. In this case, the agent again infers actions that minimize the expected free energy G, effectively maximizing information gain on the object identity. In this case, CCN votes are aggregated in the concentration parameters of a Dirichlet distribution, as described in Section~\ref{sect:method}. We also add a fixed $0.5$ value vote for the ``other'' object category which accounts for evindence for the ``other'' class when none of the CCNs ``fire''. We compare this agent with a static agent that only has a single view. In this case, only a single vote is used for the parameters of the Dirichlet distribution.

    

\begin{figure}[t!]
        \begin{subfigure}[t!]{0.72\textwidth}
            \centering
            \includegraphics[width=\textwidth]{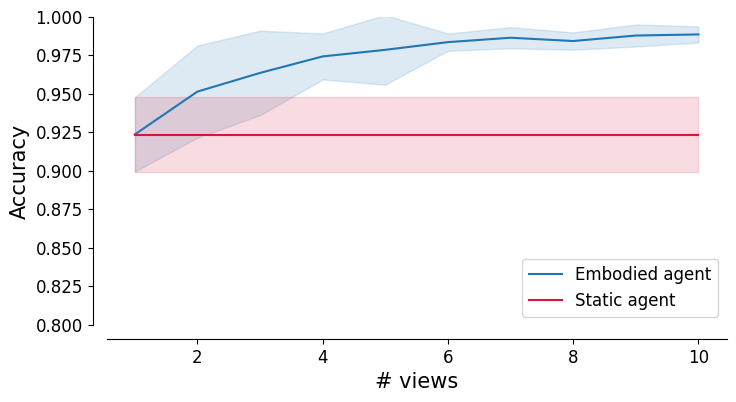}
            \caption{}
            \label{fig:embodiment}
        \end{subfigure}
        \hfill
        \begin{subfigure}[t!]{0.23\textwidth}
            \centering
            \includegraphics[width=\textwidth]{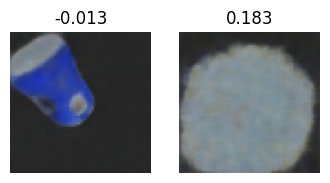}
            \includegraphics[width=\textwidth]{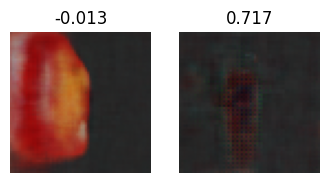}
            \includegraphics[width=\textwidth]{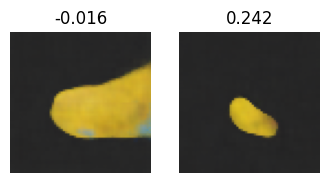}
            \caption{}
            \label{fig:fe_qualitative}
        \end{subfigure}
        \caption{(a) Performance of the CCN ensemble in a static (red) and an embodied agent, driven through active inference (blue). The agent is provided with 20 situations for each of the 9 known objects and 5 never-before seen test objects.
        (b) Imagined views for actions that result in the largest and smallest expected free energy $G$.}
    \end{figure}

    To evaluate the performance, we randomly sample 20 views for each of 14 object classes (9 known object classes and 5 never seen before), and evaluate both the static and embodied classificaton accuracy. The results are shown in Figure~\ref{fig:embodiment}. Whereas the static agent achieves an overall accuracy of 92.5\%, the embodied agent consistently improves in accuracy, reaching 98\%, as more viewpoints are queried, in line with~\cite{Hawkins2017}. The error bounds are computed over 5 different random seeds and represent the 95\% HDI. Figure~\ref{fig:fe_qualitative} shows the imagined observations for the largest and smallest expected free energy $G$. Figure~\ref{fig:fe_qualitative} shows imagined views with highest or lowest $G$, and illustrates that the active inference agent prefers observations where the object is clearly in view from a more close up view, rather than more ambiguous viewpoints.
    
\section{Related work} 

    
    Deep learning has been widely used for static image classification~\cite{alexnet,resnets}. However, recent work also focused on active vision. In~\cite{VanDeMaele} a generative model learning representations of a whole 3D scene was used for an active inference agent, whereas in \cite{Dauce20iwai} an explicit what and where stream were modeled for classifying MNIST digits. 
    
    Recently a lot of progress has been made in methods that learn the 3D geometry of objects. The geometry can either be learned implicitly using Neural Radiance Fields (NeRF)~\cite{nerf} or Generative Query Networks (GQN)~\cite{Eslami2018} or explicitly using Scene Representation Networks (SNR)~\cite{sitzmann2019srns}. However these approaches either have to generalize to a large variety of objects, which results in an involved training process requiring a lot of data, or they optimize for a single observation, limiting the flexibility. 
    
    Continual learning methods are able to use experience gathered during deployment of a system to improve the system over time. Typical approaches involve an ensemble of classifiers, that operate on a subset of the inputs, either by splitting the train data in specific subsets to train a mixture of experts~\cite{AdaptiveMixtures}, or by identifying clusters in a shared latent space and training separate classifiers for separate clusters~\cite{continuallearning}.
    
    The use of information gain has also been used as an exploration strategy outside of the active inference community, in which it substantially improves exploratory performance on a number of Atari tasks~\cite{nikolov2019informationdirected}.
    
    While most approaches tackle these problems separately, we propose a biologically inspired method that learns object-centric representations in an unsupervised manner for both the object identity and its geometric properties.
    
\section{Discussion} 

We believe that this is a first step towards manipulation of three dimensional objects, and plan to extend this work to a real-world robot setup. In this case, the robotic agent needs to make inferences on the object pose and identity that is present in the workspace. In case the object is identified, the agent is attracted to preferred observations, e.g. for grasping or manipulating the object. In case a different, novel object class is identified, a new CCN is instantiated and trained on these novel object views. In this case, we could infer the viewpoints that have a high information gain on model parameters in an active learning setting, which can also be written as an expected free energy term~\cite{Friston2016}. We can further extend the generative model to also take into account multiple objects in the scene and modelling inter-object relations and geometry. 

A limitation of our current setup is that it can only deal with a single object in the center of the view. As multi-object scenes are ubiquitous in the real world, this is a natural direction for future work. We propose a solution in which the agent can divide its spatial attention on the observations by looking at the CCN activations at different patches on the observation. Once an object and its relative reference frame is found, these can be linked using a global, ego-centric reference frame of the agent~\cite{parr2021activevision}. This way, a hierarchical generative model of the whole workspace, composed of different object is constructed. These latent parameters can then be propagated over time through a predictive model, and can in that way deal with occlusions.

In principle, one could instantiate a hierarchy of CCNs, where higher level CCNs process the output of lower level CCNs, effectively modeling part-whole relationships. This is similar to Capsule Networks~\cite{capsule} and GLOM~\cite{glom}, and corresponds better with the 1000 brains theory~\cite{Hawkings2019Cortical}. However, given the limited scalability of state of the art implementations of such hierarchical approaches~\cite{capsule}, we adopted CCNs that operate on the level that is most important for a robot operating in a workspace, i.e. the discrete object level.

We found that failure cases exist when CCNs incorrectly ``fire'' for unseen objects. This confusion occurs for some objects yielding a non-perfect classification score. We could further improve the system by also taking into account how good novel observations match our predictions in the past.

\section{Conclusion} 

In this paper we showed a novel approach to modelling 3D object properties, drawing inspiration from current development in the Neuroscientific domain. We proposed to model a separate what- and where stream for each individual object and are able to use these models for object identification as well as implicit object pose estimation. We show that through embodiment, these models can aggregate information and increase classification performance. Additionally, we show that by following the free energy formulation, these module networks can be used for implicit pose estimation of the objects. 

\subsubsection*{Acknowledgments}
This research received funding from the Flemish Government (AI Research Program). Ozan \c{C}atal was funded by a Ph.D. grant of the Flanders Research Foundation (FWO). Part of this work has been supported by Flanders Innovation \& Entrepreneurship, by way of grant agreement HBC.2020.2347.


\bibliographystyle{plain}  
\bibliography{bibliography}       

\begin{thebibliography}{10}

\bibitem{ycb}
Berk Calli, Arjun Singh, Aaron Walsman, Siddhartha Srinivasa, Pieter Abbeel,
  and Aaron~M. Dollar.
\newblock The ycb object and model set: Towards common benchmarks for
  manipulation research.
\newblock In {\em 2015 International Conference on Advanced Robotics (ICAR)},
  pages 510--517, 2015.

\bibitem{Dauce20iwai}
Emmanuel Daucé and Laurent~U Perrinet.
\newblock Visual search as active inference.
\newblock In {\em IWAI 2020}, 2020.

\bibitem{Eslami2018}
S.~M.~Ali Eslami, Danilo~Jimenez Rezende, Frederic Besse, Fabio Viola, Ari~S.
  Morcos, Marta Garnelo, Avraham Ruderman, Andrei~A. Rusu, Ivo Danihelka, Karol
  Gregor, David~P. Reichert, Lars Buesing, Theophane Weber, Oriol Vinyals, Dan
  Rosenbaum, Neil Rabinowitz, Helen King, Chloe Hillier, Matt Botvinick, Daan
  Wierstra, Koray Kavukcuoglu, and Demis Hassabis.
\newblock Neural scene representation and rendering.
\newblock {\em Science}, 360(6394):1204--1210, June 2018.

\bibitem{whatwhere}
George Ettlinger.
\newblock “object vision” and “spatial vision”: The neuropsychological
  evidence for the distinction.
\newblock {\em Cortex}, 26(3):319--341, 1990.

\bibitem{Friston2016}
Karl Friston, Thomas FitzGerald, Francesco Rigoli, Philipp Schwartenbeck, John
  O'Doherty, and Giovanni Pezzulo.
\newblock Active inference and learning.
\newblock {\em Neuroscience \& Biobehavioral Reviews}, 68:862--879, 2016.

\bibitem{adversarial}
Justin Gilmer, Ryan~P. Adams, Ian~J. Goodfellow, David~G. Andersen, and
  George~E. Dahl.
\newblock Motivating the rules of the game for adversarial example research.
\newblock {\em CoRR}, abs/1807.06732, 2018.

\bibitem{Hadsell2020}
Raia Hadsell, Dushyant Rao, Andrei~A. Rusu, and Razvan Pascanu.
\newblock Embracing change: Continual learning in deep neural networks.
\newblock {\em Trends in Cognitive Sciences}, 24(12):1028--1040, December 2020.

\bibitem{Hawkins2017}
Jeff Hawkins, Subutai Ahmad, and Yuwei Cui.
\newblock A theory of how columns in the neocortex enable learning the
  structure of the world.
\newblock {\em Frontiers in Neural Circuits}, 11:81, 2017.

\bibitem{Hawkings2019Cortical}
Jeff Hawkins, Marcus Lewis, Mirko Klukas, Scott Purdy, and Subutai Ahmad.
\newblock A framework for intelligence and cortical function based on grid
  cells in the neocortex.
\newblock {\em Frontiers in Neural Circuits}, 12:121, 2019.

\bibitem{resnets}
Kaiming He, Xiangyu Zhang, Shaoqing Ren, and Jian Sun.
\newblock Deep residual learning for image recognition.
\newblock In {\em 2016 IEEE Conference on Computer Vision and Pattern
  Recognition (CVPR)}, pages 770--778, June 2016.

\bibitem{glom}
Geoffrey~E. Hinton.
\newblock How to represent part-whole hierarchies in a neural network.
\newblock {\em CoRR}, abs/2102.12627, 2021.

\bibitem{AdaptiveMixtures}
Robert~A. Jacobs, Michael~I. Jordan, Steven~J. Nowlan, and Geoffrey~E. Hinton.
\newblock Adaptive mixtures of local experts.
\newblock {\em Neural Computation}, 3(1):79--87, 1991.

\bibitem{kingma2017adam}
Diederik~P. Kingma and Jimmy Ba.
\newblock Adam: A method for stochastic optimization, 2017.

\bibitem{kingma2013auto}
Diederik~P Kingma and Max Welling.
\newblock Auto-encoding variational bayes.
\newblock {\em arXiv preprint arXiv:1312.6114}, 2013.

\bibitem{alexnet}
Alex Krizhevsky, Ilya Sutskever, and Geoffrey~E. Hinton.
\newblock Imagenet classification with deep convolutional neural networks.
\newblock In {\em Proceedings of the 25th International Conference on Neural
  Information Processing Systems - Volume 1}, NIPS'12, page 1097–1105, Red
  Hook, NY, USA, 2012. Curran Associates Inc.

\bibitem{nerf}
Ben Mildenhall, Pratul~P Srinivasan, Matthew Tancik, Jonathan~T Barron, Ravi
  Ramamoorthi, and Ren Ng.
\newblock Nerf: Representing scenes as neural radiance fields for view
  synthesis.
\newblock In {\em European Conference on Computer Vision}, pages 405--421.
  Springer, 2020.

\bibitem{nikolov2019informationdirected}
Nikolay Nikolov, Johannes Kirschner, Felix Berkenkamp, and Andreas Krause.
\newblock Information-directed exploration for deep reinforcement learning,
  2019.

\bibitem{parr2021activevision}
Thomas Parr, Noor Sajid, Lancelot Da~Costa, M.~Berk Mirza, and Karl~J. Friston.
\newblock Generative models for active vision.
\newblock {\em Frontiers in Neurorobotics}, 15:34, 2021.

\bibitem{rezende2014stochastic}
Danilo~Jimenez Rezende, Shakir Mohamed, and Daan Wierstra.
\newblock Stochastic backpropagation and approximate inference in deep
  generative models, 2014.

\bibitem{rezende2018geco}
Danilo~Jimenez Rezende and Fabio Viola.
\newblock Taming vaes, 2018.

\bibitem{capsule}
Sara Sabour, Nicholas Frosst, and Geoffrey~E. Hinton.
\newblock Dynamic routing between capsules.
\newblock {\em CoRR}, abs/1710.09829, 2017.

\bibitem{embodied}
Adam Safron.
\newblock The radically embodied conscious cybernetic bayesian brain: From free
  energy to free will and back again.
\newblock {\em Entropy}, 23(6), 2021.

\bibitem{continuallearning}
Murray Shanahan, Christos Kaplanis, and Jovana Mitrovic.
\newblock Encoders and ensembles for task-free continual learning.
\newblock {\em CoRR}, abs/2105.13327, 2021.

\bibitem{sitzmann2019srns}
Vincent Sitzmann, Michael Zollh{\"o}fer, and Gordon Wetzstein.
\newblock Scene representation networks: Continuous 3d-structure-aware neural
  scene representations.
\newblock In {\em Advances in Neural Information Processing Systems}, 2019.

\bibitem{activeinferencetutorial}
Ryan Smith, Karl Friston, and Christopher Whyte.
\newblock A step-by-step tutorial on active inference and its application to
  empirical data, Jan 2021.

\bibitem{VanDeMaele}
Toon Van~de Maele, Tim Verbelen, Ozan Çatal, Cedric De~Boom, and Bart Dhoedt.
\newblock {\em Frontiers in Neurorobotics}, 15:14, 2021.

\end{thebibliography}

\clearpage
\appendix 

\begin{subappendices}
\renewcommand{\thesection}{\Alph{section}}

\section{Neural Network Architecture and Training Details}
\label{appendix:neuralnets}

    The neural network is based on a variational autoencoder~\cite{kingma2013auto,rezende2014stochastic} consisting of an encoder and a decoder. The encoder $\phi_\theta$ uses a convolutional pipeline to map a high dimensional input image (64x64x3) into a low dimensional latent distribution. We parameterize this distribution as a Bernouilli distribution representing the identity of the object and the camera viewpoint as a Multivariate Normal distribution with diagonal covariance matrix of 8 latent dimensions. The decoder $\psi_\theta$ then takes a sample from the viewpoint and is able to reconstruct the observation through a convolutional pipeline using transposed convolutions. In addition to a traditional variational autoencoder, we have a transition model $\chi_\theta$ that transforms a sample from the viewpoint distribution to a novel latent distribution, provided with an action. This action is a 7D vector representing the translation as coordinates in and rotation in quaternion representation. The model architecture for encoder, decoder and transition models are shown in Table~\ref{tab:encoder}, Table~\ref{tab:decoder} and Table~\ref{tab:transition}, respectively. 

    The model is optimized end-to-end through the minimization of Free Energy as described in Equation~\ref{eq:fe}. The expectations over the different terms are approximated through stochastic gradient descent using the Adam optimizer~\cite{kingma2017adam}. As minimization of negative log likelihood over reconstruction is equivalent to minimization of the Mean Squared Error, this is used in practice. Similarly, the negative log likelihood over the identity is implemented as a binary cross-entropy term. We choose the prior belief over $\mathbf{v}$ to be an isotropic Gaussian with variance 1. The individual terms of the loss function are constrained and weighted using Lagrangian multipliers~\cite{rezende2018geco}. We consider only a single timestep during the optimization process. In practice this boils down to: 
    
    \begin{equation}
        \begin{split}
            L_{FE} = &\lambda_1 \cdot L_{BCE}(\hat{i}, i) + \lambda_2 \cdot L_{MSE}(\psi_\theta(\mathbf{\hat{v}}_{t+1}), \mathbf{o}_{t+1}) \\ 
            &+ D_{KL}[ \underbrace{\chi_\theta(\mathbf{v}_t, \mathbf{a}_t)}_{q(\mathbf{v}_{t+1}|\mathbf{v}_{t}, \mathbf{a}_t, \mathbf{i})} || \underbrace{\phi_\theta(\mathbf{\hat{o}})}_{p(\mathbf{v}_{t+1}|\mathbf{i}, \mathbf{o}_t)} ]
        \end{split}
    \end{equation}

    where $\hat{i}$ is the prediction  $\phi_\theta(\mathbf{o}_t)$ of the what-stream for the encoder, $\mathbf{\hat{v}_{t+1}}$ is a sample from the predicted transitioned distribution $\chi_\theta(\mathbf{v}_t, \mathbf{a}_t)$ and $\mathbf{\hat{o}}_{t+1}$ is the expected observation from viewpoint $\mathbf{\hat{v}_{t+1}}$, decoded through $\psi_\theta(\mathbf{v}_{t+1})$. The $\lambda_{i}$ variables represent the Lagrangian multipliers used in the optimization process. 
    
    During training, pairs of observations $\mathbf{o}_t$ and $\mathbf{o}_{t+1}$ and corresponding action $\mathbf{a}_t$ are required. To maximize data efficiency, the equation is also evaluated for zero-actions using only a single observation, and reconstructing this directly without transition model. 
    
    
    

  \begin{table}[H]
      \caption{Neural network architecture for the image encoder. All strides are applied with a factor 2. The input image has a shape of 3x64x64. The output of the convolutional pipeline is used for three different heads. The first predicts the mean of the distribution $\mu$, the second head predicts the natural logarithm of the variance $\sigma^2$, for stability reasons and finally the third head predicts the classification output score $c$ as a value between zero and one after activation through the sigmoid activation function.}
      \label{tab:encoder}
      \centering
        \begin{tabular}{llcc}
            \toprule
            \textbf{Output label} & \textbf{Layer} & \textbf{Kernel size} & \textbf{\# Filters} \\
            \midrule 
            & Strided Conv2D & 4 & 8 \\
            & LeakyReLU & \\
            & Strided Conv2D & 4 & 16 \\ 
            & LeakyReLU & \\ 
            & Strided Conv 2D & 4 & 32 \\ 
            & LeakyReLU & \\ 
            & Strided Conv2D & 4 & 64 \\
            & LeakyReLU & \\ 
            $h$ & Reshape to 128 & \\
            $\mu$ & Linear (input: $h$) & & 8 \\
            ln $\sigma^2$ & Linear (input: $h$) & & 8 \\
            $c$ & Linear + Sigmoid (input: $h$) & & 1 \\
            \bottomrule 
        \end{tabular}
    \end{table}

    \begin{table}[H]
        \centering
        \caption{Neural network architecture for the image decoder. The input of this model is a sample drawn from the latent distribution, either straight from the encoder, or transitioned through the transition model. All transpose layers use a stride of two. The final layer of the model is a regular convolution with stride 1 and kernel size 1, after which a sigmoid activation is applied to map the outputs in the correct image range.}
        \label{tab:decoder}
        \begin{tabular}{lcc}
        \toprule
            \textbf{Layer} & \textbf{Kernel size} & \textbf{\# Filters}  \\
        \midrule 
            Linear & & 128 \\ 
            Reshape to 128x1x1 & & \\ 
            ConvTranspose2D & 5 & 64 \\
            LeakyReLU & \\ 
            ConvTranspose2D & 5 & 64 \\
            LeakyReLU & \\ 
            ConvTranspose2D & 6 & 32 \\ 
            LeakyReLU & \\ 
            ConvTranspose2D & 6 & 16 \\ 
            LeakyReLU \\ 
            Conv2D & 1 & 3 \\ 
            Sigmoid \\ 
        \bottomrule 
        \end{tabular}
    \end{table}
    
    \begin{table}[H]
      \centering
        \caption{Neural network architecture for the transition model. The input from this model is an 8 dimensional latent code, concatenated with the 7-dimensional representation of the relative transform (position coordinates and orientation in quaternion representation). For stability reasons, the log-variance is predicted rather than the variance directly.}
        \label{tab:transition}
        \begin{tabular}{llc}
            \toprule 
            \textbf{Output label} & \textbf{Layer} & \textbf{\# Filters} \\
            \midrule 
            & Linear & 128 \\
            & LeakyReLU & \\
            & Linear & 256 \\ 
            & LeakyReLU & \\
            & Linear & 256 \\ 
            & LeakyReLU & \\ 
            $\mu$ & Linear & 8 \\ 
            ln $\sigma^2$ & Linear & 8 \\ 
            \bottomrule
        \end{tabular}
    \end{table}
  
\section{Additional experimental details}

\begin{figure}
    \centering
    \includegraphics[width=.103\textwidth]{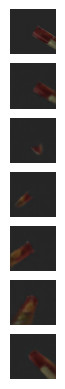}
    \includegraphics[width=.103\textwidth]{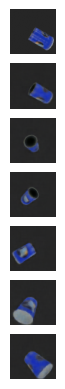}
    \includegraphics[width=.103\textwidth]{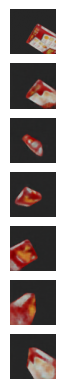}
    \includegraphics[width=.103\textwidth]{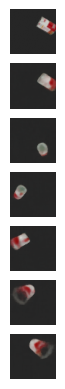}
    \includegraphics[width=.103\textwidth]{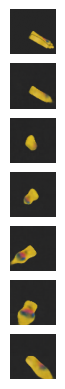}
    \includegraphics[width=.103\textwidth]{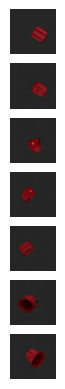}
    \includegraphics[width=.103\textwidth]{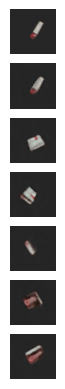}
    \includegraphics[width=.103\textwidth]{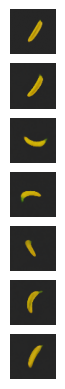}
    \includegraphics[width=.103\textwidth]{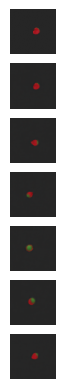}
    \caption{The top row represents the ground truth observation that was provided as input to the model. The second row shows a direct reconstruction when no action is applied to the transition model. All subsequent rows show imagined observations along a trajectory.}
    \label{fig:imaginations}
\end{figure}

In Table~\ref{tab:distance}, the computed angular and translational distances for the 9 evaluated objects are shown. Figure~\ref{fig:imaginations} shows a sequence of imaginations for all 9 objects, the top row represents the ground truth input, the second row the reconstruction and the subsequent rows are imagined observations along a trajectory. 

\begin{table}[H]
    \centering
    \caption{The mean distance error in meters and angle error in radians for different objects of the YCB dataset~\cite{ycb} in our simulated environment. For each object 20 arbitrary target poses were generated over which the mean values are computed. \label{tab:distance}}
    \begin{tabular}{lcc}
        \toprule 
        \textbf{Object} & \textbf{Distance error (m)} & \textbf{Angle error (rad)} \\ 
        \midrule
        chips can & 0.00328 $\pm$ 0.00824 & 0.15997 $\pm$ 0.21259 \\ 
        master chef can & 0.00036 $\pm$ 0.00034 & 0.06246 $\pm$ 0.03844 \\ 
        cracker box & 0.00028 $\pm$ 0.00023 &0.04659 $\pm$ 0.02674 \\ 
        tomato soup can & 0.00073 $\pm$ 0.00104 &  0.08653 $\pm$ 0.07021 \\ 
        mustard bottle & 0.00070 $\pm$ 0.00072 & 0.06351 $\pm$ 0.03818 \\ 
        mug & 0.00083 $\pm$ 0.00128 & 0.09098 $\pm$ 0.10232 \\ 
        pudding box & 0.00051 $\pm$ 0.00052 & 0.06190 $\pm$ 0.03843  \\ 
        banana & 0.00055 $\pm$ 0.00042 & 0.07482 $\pm$ 0.03592 \\ 
        strawberry & 0.00573 $\pm$ 0.01181 & 0.16699 $\pm$ 0.15705 \\ 
        \bottomrule
    \end{tabular}
\end{table}

\end{subappendices}
\end{document}